\documentclass[10pt,twocolumn,letterpaper]{article}

\usepackage{iccv}
\usepackage{times}
\usepackage{epsfig}
\usepackage{graphicx}
\usepackage{amsmath}
\usepackage{amssymb}
\usepackage{subfigure}
\usepackage{multirow}
\usepackage{multicol}

\usepackage[pagebackref=true,breaklinks=true,letterpaper=true,colorlinks,bookmarks=false]{hyperref}

\iccvfinalcopy 

\def\iccvPaperID{13} 

\ificcvfinal\fi
\begin{document}

\title{Feature Aggregation Network for Video Face Recognition}

\author{Zhaoxiang Liu\\
CloudMinds\\
{\tt\small robin.liu@cloudminds.com}
\and
Huan Hu\\
CloudMinds\\
{\tt\small hans.hu@cloudminds.com}
\and
Jinqiang Bai\\
Beihang University\\
{\tt\small baijinqiang@buaa.edu.cn}
\and
Shaohua Li\\
CloudMinds\\
{\tt\small shaohua.li@cloudminds.com}
\and
Shiguo Lian \\
CloudMinds\\
{\tt\small sg\_lian@163.com}
}

\maketitle

\begin{abstract}
  This paper aims to learn a compact representation of a video for video face recognition task. We make the following contributions: first, we propose a meta attention-based aggregation scheme which adaptively and fine-grained weighs the feature along each feature dimension among all frames to form a compact and discriminative representation. It makes the best to exploit the valuable or discriminative part of each frame to promote the performance of face recognition, without discarding or despising low quality frames as usual methods do. Second, we build a feature aggregation network comprised of a feature embedding module and a feature aggregation module. The embedding module is a convolutional neural network used to extract a feature vector from a face image, while the aggregation module consists of cascaded two meta attention blocks which adaptively aggregate the feature vectors into a single fixed-length representation. The network can deal with arbitrary number of frames, and is insensitive to frame order. Third, we validate the performance of proposed aggregation scheme. Experiments on publicly available datasets, such as YouTube face dataset and IJB-A dataset, show the effectiveness of our method, and it achieves competitive performances on both the verification and identification protocols.
\end{abstract}

\begin{figure*}
	\begin{center}
		\includegraphics[scale=0.6]{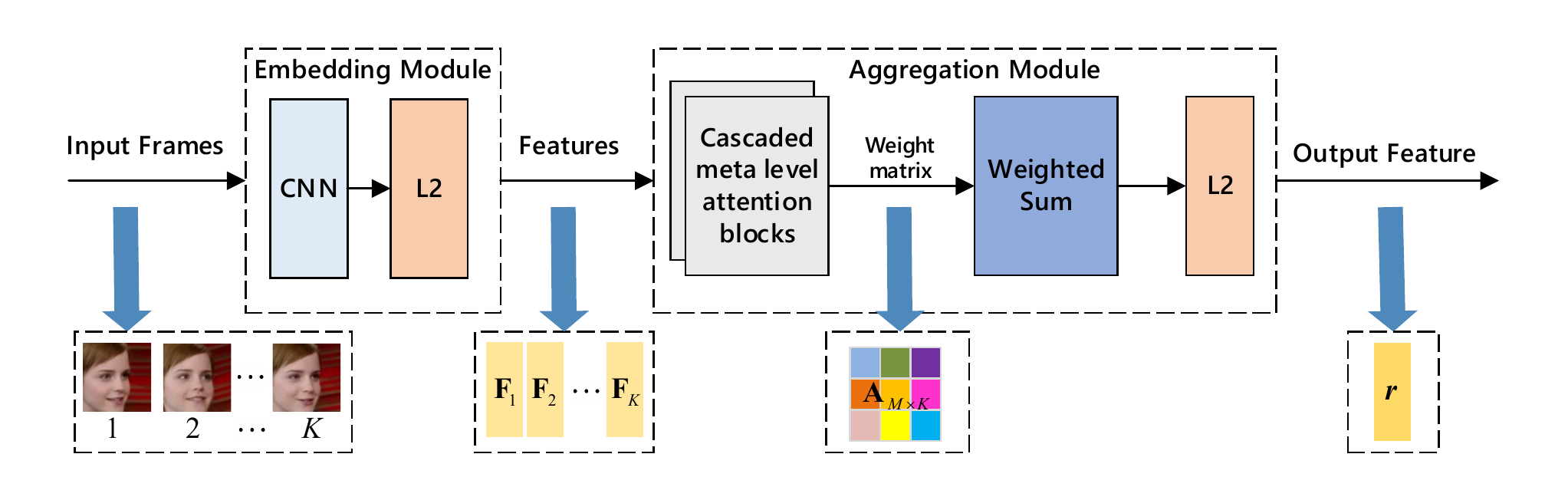}
	\end{center}
	\caption{Network architecture.  Input frames of a video are fed into feature embedding module to produce a set of normalized feature vectors. Then these features are passed through aggregation module to obtain a single fixed-size normalized feature vector for the video. The aggregation module mainly consists of cascaded two axis-level attention blocks which adaptively weighs the feature vectors along each feature axis among all frames, fusing the feature vectors organically.}
	\label{fig:1}
\end{figure*}
\section{Introduction}

Video face recognition has become more and more significant in the past few years \cite{1,2,3,4,5,6,7,8,9,10,11,12,13,14,15}, which plays an important role in many practical applications such as visual surveillance, access control, person identification, video search and so on. Compared to single still image-based face recognition, further useful information of a single face can be exploited in the video. However, the video faces exhibit much richer uncontrolled variations, \eg, out-of-focus blur, motion blur, occlusion, varied illuminations and a large range of pose variations, which make video face recognition a challenging task. Hence, how to design a feature model which can effectively represent the video face across different frames becomes a key issue of video face recognition.

In video face recognition task, each subject usually has a varied number of face images. A straightforward approach would be to represent a video face as a set of face descriptors extracted by a deep neural network, compare every pair of face descriptors between two face videos \cite{3,16}, and fuse the matching results across all pairs. However, this method would be considerably memory-consuming and inefficient especially for a large-scale recognition task. Consequently, an effective aggregation scheme, requiring minimal memory storage and supporting efficient similarity computation, is desired for this task, to generate a compact representation for a face video. And what is more, the aggregated representations should be discriminative, \ie, they are expected to have smaller intra-class distance than inter-class distance under a suitably chosen metric space.

So far, a variety of efforts on integrating information across different frames have been dedicated \cite{4,5,6,7,8,14,15,17,18,19}. Besides max pooling \cite{18}, average pooling \cite{4,5,15,18} may be the most common aggregation technique. However, it considers all frames of equal importance during feature aggregation, in which case the low quality frames with some misleading feature would degrade the performance of recognition. Considering of this problem, some other methods either just focus on high quality frames, \ie, feature-rich frames, while ignoring low quality frames, such as blurred faces, occluded faces and large pose faces \cite{6,12} or adaptively high weigh high quality frames while framesdown weigh low quality frames \cite{7,13}.

Despite that those aggregation strategies have been shown to be effective in the previous works, we believe that an optimal aggregation strategy should not simply and crudely despise the low quality frames, because the low quality frames might even contain local discriminative features which can be complementary to high quality frames. In some sense, the low quality frames may be beneficial to video face recognition. Thus, the best aggregation result should be the composition of local discriminative features from low quality frames and other parts from high quality frames. Our intuition is simple and straightforward: an ideal algorithm should be able to emphasize the valuable part of the frame feature while suppress the worthless part of the frame feature irrespective of the face quality during aggregation, \ie, it adaptively deals with each dimension of frame feature with different importance, not like NAN \cite{7} that treats each dimension of equal importance for frame feature when aggregating. Let us imagine an extreme case: with some poor quality face images, \eg, a variety of large pose faces each with different pose, it is possible to aggregate these faces into a discriminative face representation for video face recognition.

To this end, we propose a new attention-based aggregation network which adaptively and fine-grained weighs the feature along each feature dimension among all frames to form a compact and discriminative face representation. Different from previous methods, we neither focus only on high quality frames nor simply weigh the feature on frame-level. Instead, we design a neural network which is able to adaptively and fine-grained measure the importance of each dimension of the feature among all frames.

Our major contributions can be summarized as follows:
\begin{itemize}
	\item We propose a novel feature aggregation scheme for video face recognition, and reveal why it could work better. It is a generalized feature aggregation scheme, and may also serve as a feature aggregation scheme for other computer vision tasks.
	\item Based on the proposed aggregation scheme, we construct a feature aggregation network (as shown in Figure \ref{fig:1}) composed of two modules trained end-to-end or one by one separately. One is the feature embedding module which is a frame-level feature extractor using deep CNN model. The other is the aggregation module which adaptively integrates the feature vectors of all the video frames together. Our feature aggregation network inherits the main advantages of the pooling techniques (\eg, average pooling and max pooling), could handle arbitrary input size and produces order-invariant, fixed-size feature representation.
	\item We demonstrate the effectiveness of our proposed aggregation scheme in video face recognition by various comparative experiments. Trained on publicly available dataset, such as YouTube face dataset and IJB-A dataset, our method takes a lead over the baseline methods and is a competitive method compared to the state of the art methods.
\end{itemize}

\section{Related works and preliminaries}

Since our work is concerned with order-insensitive video or image set face recognition, any other methods exploiting the temporal information of video sequence will not be considered here.

Early traditional studies attempt to represent the face videos or image sets as manifolds \cite{19,20,21,22,23,24} or convex hulls \cite{25} and compute their similarities under corresponding spaces.  While those methods may work well under constrained scenarios, they are usually incapable of dealing with large face variations.

Some other methods extract the local features of frames and aggregate them across multiple frames to represent the videos \cite{4,26,27}. For example, PEP-based methods \cite{4, 26} take a part-based representation by extracting and merging LBP or SIFT descriptors, and the method in \cite{27} applies Fisher vector encoding to represent each frame by extracting RootSIFT \cite{28,29} and fuses across multiple different video frames to form a video-level representation.

These years, still image-based face recognition has gained great success thank to deep learning techniques \cite{3,11,16,30,31}. Based on this, some simple aggregation strategies are adopted in video face recognition. The methods in \cite{3} and \cite{16} utilize pairwise frame feature similarity computation and then fuse the matching results. Max- or average-pooling is used to aggregate the frame features in \cite{5, 14, 17, 18}. Though DAN \cite{6} proposes a GAN-like aggregation network which takes the video clip as input and reconstructs a single image as output to represent the video, the average pooling result of the video frames is employed to supervise the aggregation training. What is more, DAN is not suitable to tackle image set face recognition due to that a video face discriminator is used inside the GAN.

Recently, a few methods take a lead over the simple pooling techniques. The method in \cite{12} utilizes discrete wavelet transform and entropy computation to select feature-rich frames from a video sequence and learns a joint feature from them. GhostVLAD \cite{13} employs a modified NetVLAD \cite{32} layer to down weigh the contribution of low quality frames. NAN \cite{7} proposes an attention mechanism to adaptively weigh the frames, so that the contribution of low quality frames to the aggregation is down weighed. However, NAN considers each dimension of the feature vector to be of equal importance. These methods may lose some valuable information of the low quality images. This motivates us to seek a better solution in this paper.

Our work is inspired by NAN \cite{7}. However, our aggregation scheme is a more generalized strategy, can fine-grained handle the feature vector on dimension level. Now, let us review the feature aggregation scheme of NAN \cite{7}. Consider the video face recognition task on $n$ pairs of video face data $ (S^{i},y_{i})_{i=1}^{n}$, where $S^{i}$ is a face video sequence or image set with varying image number $K_{i}$, \ie, $ S^{i}=\{x_{1}^{i},x_{2}^{i},...,x_{K_{i}}^{i}\}$ in which $ x_{k}^{i},\ k=1,2,...,K_{i} $ is the $k$-th frame in the video, and $ y_{i} $ is the corresponding subject ID of $S^{i}$. Each frame $ x_{k}^{i}$ has a corresponding normalized feature representation $\boldsymbol{F}_{k}^{i}$ extracted from the feature embedding module, and the aggregated feature representation becomes
\begin{equation}\label{eqn:1}
\begin{aligned}
\boldsymbol{r}=\Sigma_{k=1}^{K_{i}}a_{k}^{i}\boldsymbol{F}_k^i,
\end{aligned}
\end{equation}where $a_k^i$ is the linear weight generated from all feature vectors of a video, it can be formulated as
\begin{equation}\label{eqn:2}
\begin{aligned}
a_k^i=\dfrac{exp(e_{k}^{i})}{\Sigma_{j=1}^{K_{i}}exp(e_{j}^{i})},
\end{aligned}
\end{equation}where $e_k^i$ is the corresponding significance yielded via dot product with a kernel filter $\boldsymbol{q}$ for each feature vector, it can be formulated as
\begin{equation}\label{eqn:3}
\begin{aligned}
e_{k}^{i}=\boldsymbol{q}^T\boldsymbol{F}_k^i.
\end{aligned}
\end{equation}

Obviously, if $a_{k}^{i}=\dfrac{1}{K_{i}}$ , Eq. (\ref{eqn:1}) degrades to average pooling strategy.

\section{Method}
\subsection{The proposed aggregation scheme}

We argue that each dimension of the feature vector shares the common weight as NAN does is not optimal. The ideal strategy should be able to adaptively weigh each dimension of feature vector separately. So we leverage a kernel matrix $\boldsymbol{Q}$ to filter the feature vector $\boldsymbol{F}_{k}^{i}$ via product, yielding a significance vector $\boldsymbol{E}_{k}^{i}$, which describes the importance of each dimension of $\boldsymbol{F}_{k}^{i}$. Assuming $\boldsymbol{F}_{k}^{i}$ is an $M$-dimension vector, then, we can formulate $\boldsymbol{Q}$ as
\begin{equation}\label{eqn:4}
\begin{aligned}
\boldsymbol{Q}={
	\left[\begin{array}{c}
\boldsymbol{q}_1^T \\
\boldsymbol{q}_2^T \\
\vdots \\
\boldsymbol{q}_M^T \\
\end{array}
\right]_{M\times{M}},}
\end{aligned}
\end{equation} and formulate $\boldsymbol{E}_k^i$ as
\begin{equation}\label{eqn:5}
\begin{aligned}
\boldsymbol{E}_k^i={
	\left[\begin{array}{c}
	e_{1k}^i \\
	e_{2k}^i \\
	\vdots \\
	e_{Mk}^i \\
	\end{array}
	\right]_{M\times{1}}}=\boldsymbol{Q}\boldsymbol{F}_k^i={
	\left[\begin{array}{c}
	\boldsymbol{q}_1^T\boldsymbol{F}_{k}^i \\
	\boldsymbol{q}_2^T\boldsymbol{F}_{k}^i \\
	\vdots \\
	\boldsymbol{q}_M^T\boldsymbol{F}_{k}^i \\
	\end{array}
	\right]_{M\times{1}}}.
\end{aligned}
\end{equation}

\begin{figure}[b]
	\begin{center}
		\includegraphics[scale=0.4]{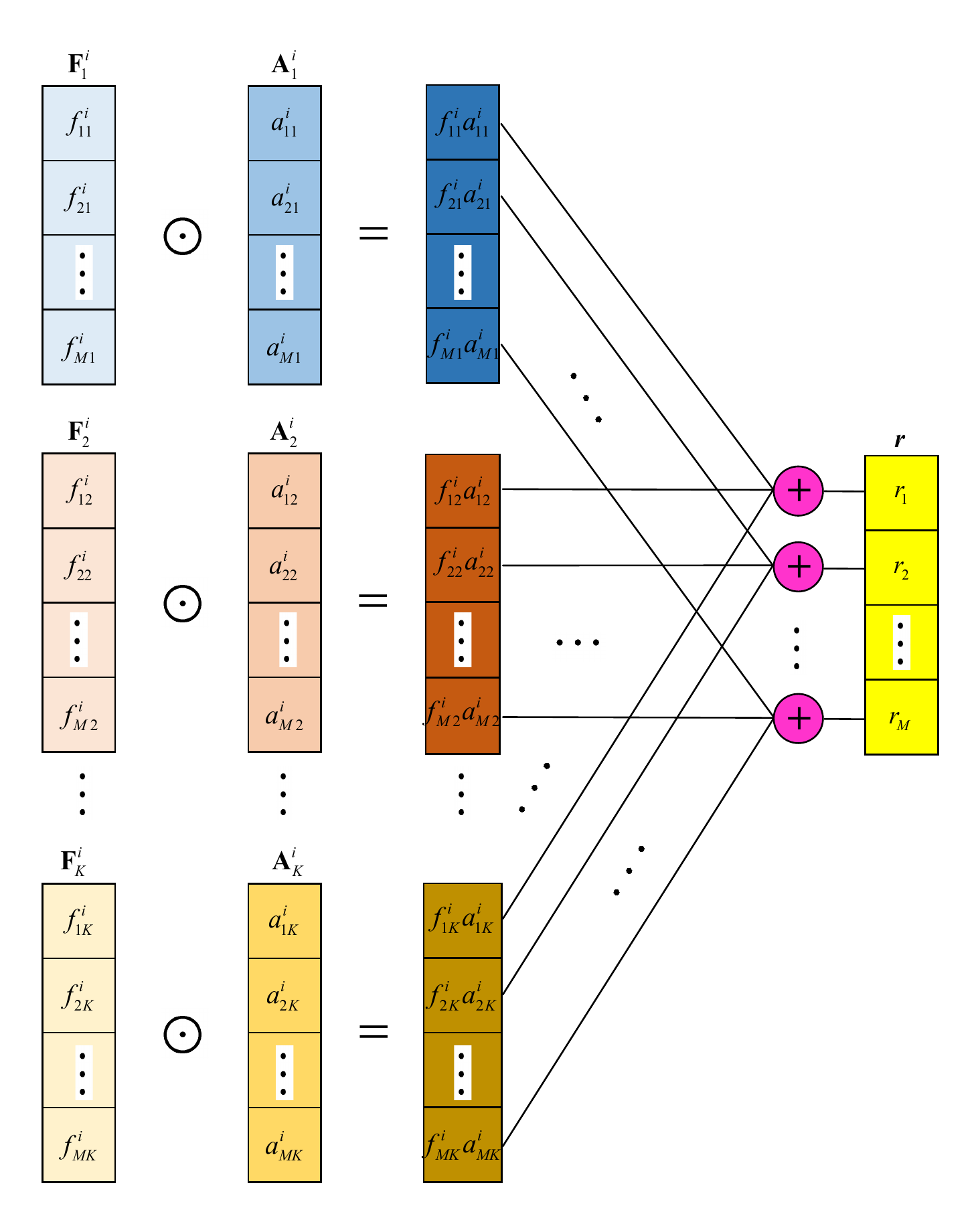}
	\end{center}
	\caption{Element-wise weighted sum of features.}
	\label{fig:2}
\end{figure}

After softmax operation along each dimension, a positive weight vector $\boldsymbol{A}_k^i$ is generated as following
\begin{equation}\label{eqn:6}
\begin{aligned}
\boldsymbol{A}_k^i={
	\left[\begin{array}{c}
	a_{1k}^i \\
	a_{2k}^i \\
	\vdots \\
	a_{Mk}^i \\
	\end{array}
	\right]_{M\times{1}}}={
	\left[\begin{array}{c}
	\dfrac{exp(e_{1k}^{i})}{\Sigma_{j=1}^{K_{i}}exp(e_{1j}^{i})} \\
	\dfrac{exp(e_{2k}^{i})}{\Sigma_{j=1}^{K_{i}}exp(e_{2j}^{i})} \\
	\vdots \\
	\dfrac{exp(e_{Mk}^{i})}{\Sigma_{j=1}^{K_{i}}exp(e_{Mj}^{i})} \\
	\end{array}
	\right]_{M\times{1}}},
\end{aligned}
\end{equation} where $a_{mk}^i$ denotes the linear weight of that $m$-th dimension of the feature vector contributes to aggregation result, and $\Sigma_{k=1}^{K_{i}}a_{mk}^{i}=1, \forall m \in \{ 1,2,...,M \}$. So that the aggregated feature representation becomes
\begin{equation}\label{eqn:7}
\begin{aligned}
\boldsymbol{r}=\Sigma_{k=1}^{K_{i}}\boldsymbol{A}_{k}^{i}\bigodot\boldsymbol{F}_k^i,
\end{aligned}
\end{equation}where $\bigodot$ represents element-wise product.  Figure \ref{fig:2} shows the calculating process of $\boldsymbol{r}$. $\boldsymbol{r}$ turns out to be $\tilde{\boldsymbol{r}}$ after $L2$-normalization. Either cosine or $L2$ distance can be used to compute the similarity.

From above formulas and Figure \ref{fig:2}, we can clearly see the difference between our method and NAN is that we use a kernel matrix instead of a kernel vector to adaptively weigh the feature. Therefore, we can measure the importance of feature on dimension level without constraining each dimension to share the same weight just as NAN \cite{7} does. Compared to NAN and other pooling techniques, our method is more flexible, and can make each dimension of one feature vector adaptively contribute to the aggregation feature. In theory, it can realize optimal feature aggregation after well trained. So, our method can deal with every frame fairly regardless of face quality, and make the best to exploit its any valuable or discriminative local feature to promote the video face recognition.

What is more, our method is a more generalized feature aggregation scheme. Obviously, if $a_{1k}^i=a_{2k}^i=\cdots=a_{Mk}^i$, Eq. (\ref{eqn:7}) degrades to NAN, and if $a_{mk}^i=\dfrac{1}{K_i}$, Eq. (\ref{eqn:7}) degrades to average pooling. And max pooling can also be regarded as a special case of our method.
\subsection{The proposed feature aggregation network}
Based the on proposed aggregation scheme, we construct a feature aggregation network comprised of two modules. As shown in Figure \ref{fig:1}, the network can be fed with a set of face images of a subject and produces a single feature vector as its representation for the recognition task. It is built upon a modern deep CNN model for frame feature embedding, and adaptively aggregates all frames in the video into a compact vector representation.

The image embedding module of our network adopts the backbone network of Arc-Face \cite{30} which greatly advances the image-based face recognition recently. The embedding module mainly consists of a ResNet50 which has an improved residual unit: BN-Conv-BN-PReLu-Conv-BN structure, while using BN-Dropout-FC-BN after the last convolutional layer. The embedding module produces 512-dimension image features which are first normalized to be unit vectors then fed into the aggregation module.

\begin{figure}[htbp]
	\begin{center}
		\includegraphics[scale=0.6]{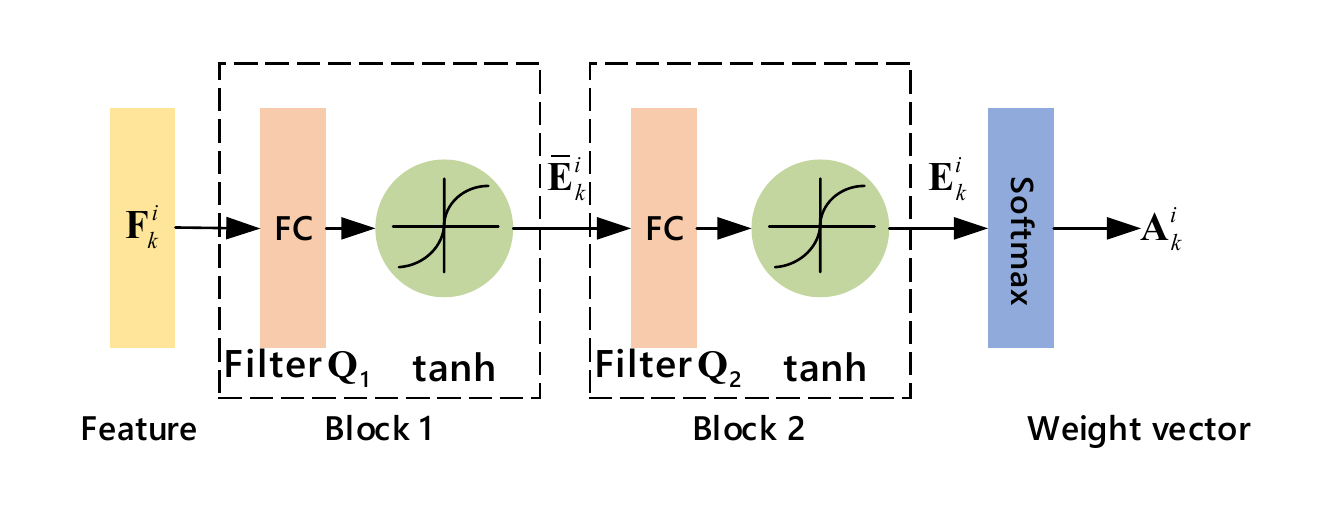}
	\end{center}
	\caption{Cascaded two attention blocks.}
	\label{fig:3}
\end{figure}

In order to obtain a better aggregation representation, a cascaded two attention blocks with nonlinear transfer is designed inside aggregation module as shown in Figure \ref{fig:3}. Each attention block consists of a kernel filter and a nonlinear transfer. The kernel filter is implemented with a $FC$ layer, while nonlinear transfer with a $\tanh$ activation layer. Then $\boldsymbol{E}_k^i$ becomes
\begin{equation}\label{eqn:8}
\begin{aligned}
\boldsymbol{E}_k^i=\tanh(\boldsymbol{Q}_2\overline{\boldsymbol{E}}_{k}^i+\boldsymbol{b}_2),
\end{aligned}
\end{equation}where $\overline{\boldsymbol{E}}_{k}^i$ is the output of the first block, it can be formulated as
\begin{equation}\label{eqn:9}
\begin{aligned}
\overline{\boldsymbol{E}}_k^i=\tanh(\boldsymbol{Q}_1\boldsymbol{F}_{k}^i+\boldsymbol{b}_1).
\end{aligned}
\end{equation}

Therefore, besides kernel matrices $\boldsymbol{Q}_1$ and $\boldsymbol{Q}_2$, biases $\boldsymbol{b}_1$ and $\boldsymbol{b}_2$ are also trainable parameters of aggregation module. We have to point out that our cascaded attention blocks are totally different from NAN \cite{7}'s in that our attention block uses an importance matrix while NAN uses an importance vector to weigh the feature vectors. In comparison, our method is more fine-grained than NAN to aggregate feature vectors. Furthermore, NAN aggregates the feature vectors twice, where the second attention block takes the aggregation result of the first attention block as input. However, our method only makes aggregation once.

In addition, our network has several other favorable properties. First, it is able to tackle arbitrary number of images for one subject. Second, the aggregation result $\boldsymbol{r}$ which is of the same size as a single feature $\boldsymbol{F}_k^i$ is invariant to the image order, keeps unchanged when the image sequence are reshuffled or even reversed, \ie, our network is insensitive to the temporal information of the video or image set. Third, it is adaptive to the input faces and whose all parameters are trainable through supervised learning with standard backpropagation and gradient descent.

\subsection{Network training}
To make the training faster and more stable, we divide it into three stages(as shown in Fig.4). Firstly, we train the embedding module for single image face recognition task. In this stage, the cleaned MS-Celeb-1M dataset \cite{30,39} is used. Secondly, we train the whole network end-to-end for set-based face recognition task, and the VGGFace2 dataset \cite{45} is used in this stage. In order to boost the capability of handling images of varying quality that typically occur in the wild, the VGGFace2 datatset is augmented in the form of image degradation, such as blurring or compression. Finally, we finetune the whole network end-to-end on the training set of the benchmark dataset.

\begin{figure}[htbp]
	\begin{center}
		\includegraphics[scale=0.25]{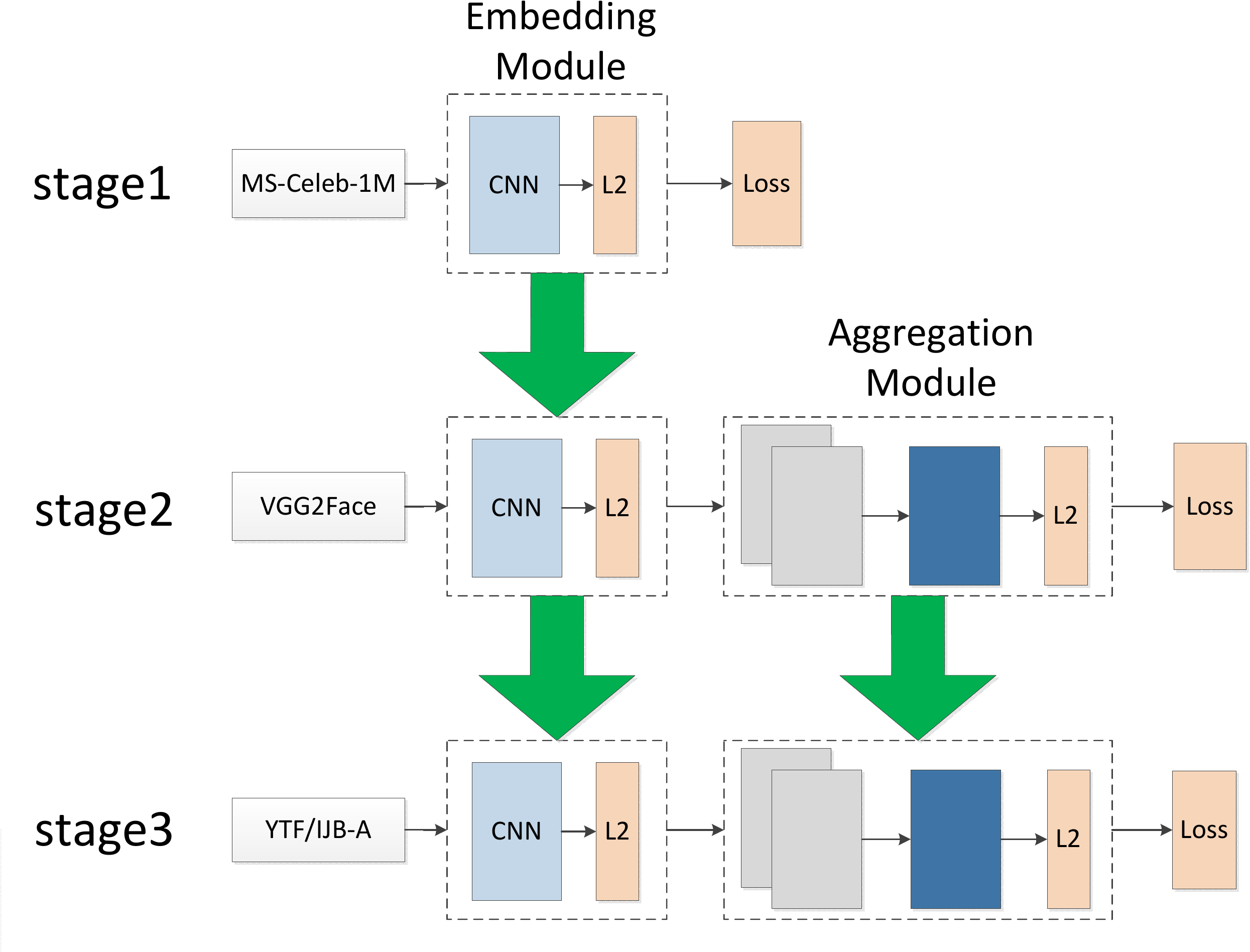}
	\end{center}
	\caption{Network Training. In stage 1, only embedding module is trained; then the trained embedding module is copied to stage 2 for end-to-end training; finally, the whole network is copied to stage 3 for end-to-end finetuning. }
	\label{fig:4}
\end{figure}

\section{Experiments}
\subsection{Datasets and protocols}
We conduct experiments on two widely used datasets including the YouTube Face dataset (YTF) \cite{36}, IJB-A dataset \cite{37}. In this section, we will first introduce our implementation details, and then report the performance of our method on above two datasets.

\subsection{Training details}
\textbf{Embedding module training:} As aforementioned, the cleaned MS-Celeb-1M dataset \cite{30,39} which contains about 3.8M images of 85k unique identities is used to train our feature embedding network for the single image face recognition task. MTCNN \cite{40} is employed to detect 5 facial landmarks in the face images. The faces are aligned to $112\times112$ by using similarity transformation according to the landmarks detected, and then fed into embedding network for training. The Additive Angular Margin Loss \cite{30}, which is a kind of modified softmax loss is used to supervise the training. After training, the classification loss layer is removed from the trained network. The rest network is fixed and used to extract a single fixed-size representation for the face image.

\textbf{End-to-End training:} We use the VGGFace2 dataset \cite{45} to train the whole network end-to-end for the set-based face recognition task. The VGGFace2 Dataset \cite{45} consists of about 3 million images, covering 8631 identities, and there are on average 360 face images for each identity. To perform set-based face recognition training, the image sets are built by repeatedly sampling a fixed number of images which belong to the same identity. All the images sampled are aligned using the same way as in the embedding module training. After alignment, the data augmentation is performed by image degradation. Following the same strategy as in GhostVLAD \cite{13},four methods: isotropic blur, motion blur, decreased resolution and JPEG compression are adopted to degrade the face image for training. The Additive Angular Margin Loss \cite{30}is also adopted to supervise the end-to-end training. In order to speed up the training, we initialize all the parameters of the aggregation module to be zero. That means the aggregation module begins with average pooling to search the optimal parameters.

\textbf{Finetuning:} All the video face dataset are also aligned by using MTCNN \cite{40} algorithm and similarity transformation.Then the whole network is finetuned on the training set of each video face dataset using the Additive Angular Margin Loss \cite{30}.

\begin{table}[b]
	\begin{center}
		\begin{tabular}{|c|c|c|}
			\hline
			Method & Accuracy(\%) & AUC \\
			\hline
			EigenPEP\cite{4} & 84.80 $\pm$ 1.40 &92.60\\
			DeepFace-single\cite{16} & 91.40 $\pm$ 1.1 &96.30\\
			DeepID2+\cite{43} & 93.20 $\pm$ 0.20 &92.30\\
			FaceNet\cite{3} & 95.12 $\pm$ 0.39 &92.30\\
			Wen et al.\cite{11} & 94.90 &92.30\\
			TBE-CNN\cite{14} & 94.96 $\pm$ 0.31 &92.30\\
			NAN\cite{7} & 95.72 $\pm$ 0.64 &98.80\\
			ADRL\cite{44} & 96.52 $\pm$ 0.54 &-\\
			Deep FR\cite{5} & 97.30 &92.30\\
			\hline
			AvgPool & 95.70 $\pm$ 0.61 &98.69\\
			NAN* & 95.93 $\pm$ 0.62 &98.92\\
			\hline
			Ours & \textbf{96.21} $\pm$ \textbf{0.63} &\textbf{99.1}\\
			\hline
		\end{tabular}
	\end{center}
	\caption{Performace evaluation on YTF benchmark. (NAN* represents the NAN \cite{7} method we reproduce with our embedding module.)\label{tab4}}
\end{table}

\subsection{Baseline methods}
Since average pooling is a widely used aggregation method in many previous works \cite{4,5,15,18}, we choose average pooling as one of our baselines. For fairness, the average pooling method shares common embedding module with our method after the whole network is finetuned on each benchmark dataset. We also choose NAN \cite{7} as our another baseline. We reproduce the NAN which consists of cascaded two attention blocks as \cite{7} describes. The reproduced NAN is trained in the same way as our method. The two baselines as well as our method produce 512-d feature representation for each video and compute the similarity in $O(1)$ time. Besides the above two baselines, we also compare with some other sate-of-the-art methods.


\begin{table*}[htbp]
	\small
	\begin{center}
		\renewcommand\tabcolsep{11.5mm}
		\begin{tabular}{|c|c|c|c|}
			\hline
			\multirow{2}{*}{Method} & \multicolumn{3}{c|}{1:1 verification TAR(\%)}\\
			\cline{2-4} & FAR=0.001 & FAR=0.01 & FAR=0.1\\
			\hline
			DREAM\cite{47} &  86.8 $\pm$ 1.5 & 94.4 $\pm$ 0.9 & - \\
			Triplet Embedding\cite{48} &  81.3 $\pm$ 2 & 91 $\pm$ 1 & 96.4 $\pm$ 0.5\\
			Template Adaptation\cite{49} & 83.6 $\pm$ 2.7 & 93.9 $\pm$ 1.3 & 97.9 $\pm$ 0.4\\
			NAN\cite{7} &  88.1 $\pm$ 1.1 & 94.1 $\pm$ 0.8 & 97.8 $\pm$ 0.3\\
			QAN\cite{50} &  89.31 $\pm$ 3.92 & 94.20 $\pm$ 1.53 & 98.02 $\pm$ 0.55\\
			VGGFace2\cite{45} &  92.1 $\pm$ 1.4 & 96.8 $\pm$ 0.6 & 99.0 $\pm$ 0.2\\
			GhostVLAD\cite{13} &  93.5 $\pm$ 1.5 & 97.2 $\pm$ 0.3 & \textbf{99.0} $\pm$ \textbf{0.2} \\
            Ranjan \etal~\cite{51} &  \textbf{95.2}  & 96.9  & 98.4 \\
			\hline
			AvgPool &  88.82 $\pm$ 1.22 & 96.18 $\pm$ 0.92 & 98.16 $\pm$ 0.40\\
			NAN* &  93.12 $\pm$ 1.16 & 96.91 $\pm$ 0.83 & 98.71 $\pm$ 0.599\\
			\hline
			Ours  &  93.61 $\pm$ 1.51 & \textbf{97.28} $\pm$ \textbf{0.28} & 98.94 $\pm$ 0.31\\
			\hline
		\end{tabular}
	\end{center}
	\caption{Performance evaluation for verification on IJB-A benchmark. The true accept rates (TAR) vs. false positive rates (FAR) are reported. (NAN* represents the NAN \cite{7} method we reproduce with our embedding module.)\label{tab2}}
\end{table*}

\begin{table*}[htbp]
	\small
	\begin{center}
		\renewcommand\tabcolsep{5mm}
		\begin{tabular}{|c|c|c|c|c|c|c|c|c|}
			\hline
			\multirow{2}{*}{Method} & \multicolumn{5}{c|}{1:N identification TPIR(\%)}\\
			\cline{2-6} & FPIR=0.01 & FPIR=0.1 & Rank-1 & Rank-5 & Rank-10\\
			\hline
			DREAM\cite{47} & - & - & 94.6 $\pm$ 1.1 & 96.8 $\pm$ 1.0 & -\\
			Triplet Embedding\cite{48} & 75.3 $\pm$ 3 & 86.3 $\pm$ 1.4 & 93.2 $\pm$ 1 & - & 97.7 $\pm$ 0.5\\
			Template Adaptation\cite{49} & 77.4 $\pm$ 4.9 & 88.2 $\pm$ 1.6 & 92.8 $\pm$ 1.0 & 97.7 $\pm$ 0.4 & 98.6 $\pm$ 0.3\\
			NAN\cite{7} & 81.7 $\pm$ 4.1 & 91.9 $\pm$ 0.9 & 95.8 $\pm$ 0.5 & 98.0 $\pm$ 0.5 & 98.6 $\pm$ 0.3\\
			VGGFace2\cite{45} & 88.3 $\pm$ 3.8 & 94.6 $\pm$ 0.4 & 98.2 $\pm$ 0.4 & 99.3 $\pm$ 0.2 & 99.4 $\pm$ 0.1\\
			GhostVLAD\cite{13} & 88.4 $\pm$ 5.9 & 95.1 $\pm$ 0.5 & 97.7 $\pm$ 0.4 & 99.1 $\pm$ 0.3 & \textbf{99.4} $\pm$ \textbf{0.2}\\
            Ranjan \etal~\cite{51} &  \textbf{92.0}  & \textbf{96.2}  & 97.5  & 98.6 & 98.9 \\
			\hline
			AvgPool & 86.43 $\pm$ 4.81 & 94.05 $\pm$ 1.02 &95.69 $\pm$ 0.62 &98.52 $\pm$ 0.45  &99.04 $\pm$ 0.33 \\
			NAN* & 87.92 $\pm$ 5.44 & 94.83 $\pm$ 1.01 &97.23 $\pm$ 0.57 &99.05 $\pm$ 0.58  &99.24 $\pm$ 0.44 \\
			\hline
			Ours  & 88.51 $\pm$ 5.86 & 95.18 $\pm$ 1.02 &\textbf{97.92} $\pm$ \textbf{0.32} &\textbf{99.23} $\pm$ \textbf{0.36}  &99.39 $\pm$ 0.25 \\
			\hline
		\end{tabular}
	\end{center}
	\caption{Performance evaluation for identification on IJB-A benchmark. The true positive identification rate (TPIR) vs. false positive identification rate (FPIR) and the Rank-N accuracies are presented.(NAN* represents the NAN \cite{7} method we reproduce with our embedding module.)\label{tab3}}
\end{table*}

\subsection{Results on YouTube Face Dataset}
We first evaluate our method on the YouTube Face Dataset \cite{36} which contains 3425 videos of 2595 different subjects. The lengths of videos vary from 48 to 6070 frames and the average length is 181.3 frames per video. The dataset is splitted into 10 folds, and each fold consists of 250 positive (intra-subject) pairs and 250 negative (inter-subject) pairs. We follow the standard verification protocol to test our method.

Table \ref{tab4} shows the results of our method, the baseline and some other state of the art methods.
We can see that our method outperforms the two baselines, reducing the error of the best-performing baseline:NAN* by 6.88\%. This can be regarded as a proof of the effectiveness of our method. Our method also performs better than all the other state-of-the-art methods (including the original NAN \cite{7} )except the deep FR methods and ADRL method \cite{44}. The reason is that, the deep FR method benefits a lot from front face selection and triplet loss embedding with carefully selected triplets, and ADRL method \cite{44} benefits from exploiting the temporal information from the video sequence. Compared to deep FR method, our aggregation method is more straightforward and elegant without hand-crafted rules. And compared to ADRL method \cite{44}, our method is order-invariant, can be used in more potential scenarios. It is noteworthy that our reproduced NAN also performs better than original NAN \cite{7}. That is because both the embedding module and aggregation module of the reproduced NAN is trained end-to-end instead of separately, and compared to separate training, more training data is used during the end-to-end training stage.

\subsection{Results on IJB-A Dataset}\label{sec4.5}
The IJB-A Dataset \cite{37} contains 5712 images and 2085 videos, covering 500 subjects in total. The average numbers of images and videos per subject are 11.4 images and 4.2 videos. This dataset is more challenging than YouTube Face dataset \cite{36} due to it covers large range of pose variations and different kinds of image conditions.

We follow the standard benchmark procedure for IJB-A to evaluate our method on both the `compare' protocol for $1:1$ face verification and the `search' protocol for $1:N$ face identification. The true accept rates (TAR) vs. false positive rates (FAR) are reported for verification, while the true positive identification rates (TPIR) vs. false positive identification rates (FPIR) and the Rank-N accuracies are reported for identification. Table \ref{tab2} and Table \ref{tab3} show the evaluation results of different methods for verification task and identification task respectively.

From above two tables, we can see that our method outperforms the two baselines by appreciable margins in both verification task and identification task, especially reducing the error of best-performing baseline by 7.12\%,11.97\% and 17.83\% at FAR=0.001, FAR=0.01 and FAR=0.1 respectively in verification task. These solidly prove the effectiveness of our method. Compared to all the state-of-the-art methods except for Ranjan \etal~\cite{51}, our method performs a little better at FAR=0.001 and FAR=0.01, performs on par with them at FAR=0.01 where TAR values have almost saturated to a 99\% mark, and beats all of them except on Rank-10 metric where our method is on par with them and the TPIR values have saturated to a 99.4\% mark. Though Ranjan \etal~\cite{51} performed better than our method in three of all the eight metrics, they fused two deeper networks for feature embedding. One is ResNet101, the other is Inception ResNet-v2 that has 244 convolution layers, both of which are much deeper than our backbone ResNet50. What is more, they used more diverse datasets than us to train the feature embedding module. Not only the still image dataset but also other additional video dataset which is beneficial to video face recognition were utilized to train the networks. Besides, the reproduced NAN also outperforms the original NAN \cite{7} as on YTF benchmark. It is noteworthy that the gap between our method and the original NAN \cite{7} on IJB-A dataset is larger compared to the results on YTF dataset. This is because the face variations in IJB-A dataset is much larger than in YTF dataset, our method can extract more beneficial information for video face recognition.

\section{Conclusion}
We introduced a new feature aggregation network for video face recognition. Our network can adaptively and fine-grained weigh the input frames along each dimension of the feature vector and fuse them organically into a compact representation which is invariant to the frame order. Our aggregation scheme can make the best to exploit any valuable part of the features regardless of the frame quality to promote the performance of video face recognition. Experiments on YTF and IJB-A benchmarks show our method is a competitive aggregation method.

{\small
\bibliographystyle{ieee}
\bibliography{egbib}
}

\end{document}